\documentclass{article}
\usepackage[utf8]{inputenc}
\usepackage{amssymb}
\usepackage{scalerel}
\usepackage{graphicx}
\graphicspath{ {./images/} }

\title{Gradient Kernel Regression}
\author{Matt Calder}
\date{March 2021}

\begin{document}
\maketitle

\section*{Introduction}

There have been a number of papers written recently on the so called neural tangent kernel \cite{domingos2020model, hayou2021meanfield}. This work is mathematical in nature and while interesting seems to overlook the practical aspects of the gradient kernel \footnote{Here the neural tangent kernel is simply referred to as the gradient kernel, since there is no specific requirement that the underlying model be a neural network.} In what follows, practical examples of using the gradient kernel are demonstrated, a general algorithm is described, and other possible uses are discussed.

\section*{Gradient Kernel}

In \cite{domingos2020model} the authors introduce what they term the \textit{path kernel} in the context of supervised learning via deep neural networks trained using stochastic gradient descent. The path kernel measures the similarity of two input points, $x_i$ and $x_j$ as,
$$
K^{\scaleto{\textrm{path}}{4pt}}(x_i, x_j) = \int_{c(t)} \partial_{w_t} f(x_i) \cdot \partial_{w_t} f(x_j) dt
$$
where $c(t)$ is the path taken by the parameters $w$ of the network $f$ during gradient descent. The path kernel is a special case of the \textit{neural tangent kernel} \cite{jacot2020neural}, 
$$
K^{\scaleto{\textrm{ntk}}{4pt}}(x_i, x_j) = \sum _{p=1}^{P} \partial_{w_p} f(x_i) \cdot \partial_{w_p} f(x_j)
$$
where the parameters over which the sum occurs, $\{w_p\}$, are left unspecified. In both cases and in related work \cite{domingos2020model,jacot2020neural,hayou2021meanfield} the authors derive results and prove theorems but spend little attention on practical applications. Here, some of those practical applications are examined.

Given a data set $\{x_i, y_i\}$ for $i \in 1 \ldots n$, and a model $f_w(x)$ parameterized by $w$ (for example a neural network), the \textit{gradient kernel} at $w$, $x_i$, $x_j$ is calculated as,
\begin{equation}
\label{Kdef}
K_w(x_i, x_j) = \partial_w f(x_i) \cdot \partial_w f(x_j).
\end{equation}
This is simply the  neural tangent kernel evaluated at a single value of the parameter vector $w$. Note that here we are working with uni-variate valued functions $f$ (eg. a binary classifier), however, it is straight forward to extend the methods described to multi-variate $f$.

\section*{Gradient Kernel Regression}

Once in possession of a kernel function it is immediately tempting to use that kernel function in a regression context. That is, to fit a function of the form,
$$
g(x) = \sum_{i=1}^{n} \alpha_i K_w(x, x_i)
$$
to a data set. However, there are a few practical questions that need to be addressed.

Modern neural network models can have data sets with millions of examples. Forming the full kernel matrix requires $\mathcal{O}(n^2)$ space to store this matrix. A well known method to reduce these
requirements is to select a much smaller subset of examples to serve as basis examples. That will be the approach taken here. The selection of the basis examples is an interesting and deep research topic itself, but that will be ignored and the basis examples will simply be selected at random from the set of training examples.

In what follows, we are interested in examining how a kernel regression using the gradient kernel performs as the parameters $w$ are modified by gradient descent applied to the network. Unfortunately, gradient descent drives the gradient kernel into an ill-conditioned and numerically unstable regime. To combat this, the gradient kernel is not used directly, but rather it is normalized into cosine similarity form,
$$
\tilde{K}(x_i, x_j) = \frac{K_w(x_i,x_j)}{\sqrt{K_w(x_i,x_i)K_w(x_j,x_j)}}
$$
which has much better numerical properties.

\textit{Gradient kernel regression} is performed and evaluated as follows. Given a data set of example points $\{x_i, y_i\}$, a function $f_w(x)$ parameterized by $w$ we separate the examples into a training set and a testing set. A set of basis examples are chosen from the training set. The linear regression model, 
$$
g_\alpha(x) = \sum_{i \in I_{\scaleto{\textrm{basis}}{4pt}}} \alpha_i \tilde{K}_w(x, x_i)
$$
is fit by least squares to the training data,
$$ 
\min_{\alpha} \sum_{i \in  I_{\scaleto{\textrm{train}}{4pt}}}
   \left( y_i - g_\alpha(x_i) \right)^2
$$
and its error is measured on the test data,
$$
E_w = \sum_{i \in  I_{\scaleto{\textrm{test}}{4pt}}}
   \left( y_i - g_\alpha(x_i) \right)^2.
$$
Here, the dependence of the error on the underlying model parameters $w$ is made explicit, since it is that relationship that is of most interest.

\section*{MNIST Example}

To study the performance of the gradient kernel regression the well known MNIST data set was used \cite{mnist-2010}. The MNIST data consists of images of hand written digits, and a class label for each image.

The underlying neural network model used was taken from the examples provided in the PyTorch machine learning library \cite{NEURIPS2019_9015}. It consists of 2 convolutional layers and 2 fully connected layers. As given, this model produces a 10-dimensional output corresponding to the multi-class digit classification problem. For simplicity, this model was changed to a binary classifier. The data chosen was for two digits ("1", and "7"), and the classification task was to distinguish a 1 from a 7. 

For the experiments, 1000 training examples, and 1000 testing examples were selected. Both sets were balanced with 500 1's and 500 7's in each. The basis examples were chosen at random from the training set, and these too were balanced with 50 1's and 50 7's. The network was initialized with random values using default PyTorch settings.

Repeated epochs of gradient descent were performed. Each epoch corresponds to a complete pass through the training data, in this case that involved 10 gradient descent steps driven by a random batch of 100
training examples. For each epoch a linear regression was fit to the training data and tested on the testing data as described in the previous section. The test result was summarized through and accuracy score which assigned class "1" to the examples where the regression function was greater than 0.5 and assigned the class "7" to examples where the regression function was less than 0.5. 

\begin{figure}[ht]
\centering
\includegraphics[width=0.9\textwidth]{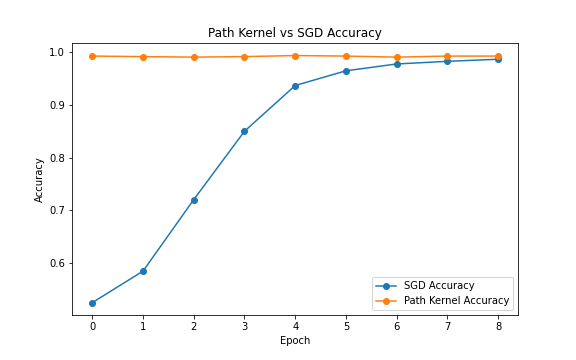}
\caption{Accuracy of gradient kernel regression vs SGD by epoch on MNIST classifier.}
\label{mnist}
\end{figure}

Figure \ref{mnist} shows the result. Surprisingly, the accuracy of the gradient kernel regression is independent of the accuracy of the underlying network. The underlying network starts with random parameters and so has a random 50\% accuracy at the start. But the gradient kernel regression is already working at the full 99\% accuracy at the start. The underlying network gets progressively better as the gradient descent epochs proceed, but at its best it only matches the level of accuracy that the gradient kernel regression holds over the whole path. 

The accuracy of the gradient kernel regression does not depend on the quality of the underlying model parameters. It works as well for random parameter settings as it does for trained parameter settings. 

\section*{CIFAR10 Example}

The result from the previous section shows that gradient kernel regression can serve as a powerful
tool in designing a network architecture. It can reveal the accuracy inherent in a network without requiring training and all the time, effort, and uncertainty that entails. 

For example, transfer learning is a method that takes an existing model and uses it as the starting point for a new model. Specifically, a large complex deep neural network trained on millions of examples can be modified by replacing its final layer with a layer customized to a new problem. Then only this new layer is trained on some smaller set of data. Gradient kernel regression can be used to explore the possible forms of this final layer efficiently. 

Here transfer learning is applied to the ResNet-50 \cite{7780459} deep neural network (as provided in PyTorch). The final layer of the ResNet-50 network is replaced with two fully connected layers and the final output is modified to be a binary classifier.

This modified network is trained to classify bird versus cat images from the CIFAR10 \cite{CIFAR} data set of image (also provided in PyTorch). Transfer learning provides an interesting use case for
the gradient kernel regression method. The gradient for very wide / deep networks can involve a huge number of parameters (ResNet-50 has over 20 million). Taking inner-products across this many parameters for each training example is computationally expensive (although, it must be noted, much less than SGD requires). Transfer learning only trains the additional layers added to the network, for the example here just over 1 million parameters.  

The experimental setup is the same as for the MNIST example. 1000 training, and 1000 testing examples were randomly selected from the CIFAR10 data. Both data sets had a balanced 500 bird and 500 cat images. 100 training examples were taken as basis examples, also balanced with 50 bird and 50 cat images. Again, 9 SGD epochs were performed and the kernels were constructed at each epoch. 

\begin{figure}[h]
\centering
\includegraphics[width=0.9\textwidth]{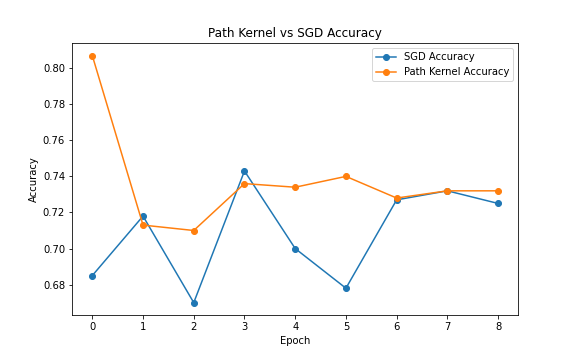}
\caption{Accuracy of path kernel vs SGD by epoch on CIFAR10 transfer classifier.}
\label{cifar10}
\end{figure}

Figure \ref{cifar10} shows the accuracy for the kernel regressions and the neural network by epoch. As before the kernel regression dominates the performance of the neural network. Perhaps most surprising, the randomly initialized model has the overall best performance, and training actually degrades the kernel result. 

\section*{Conclusion}

The examples presented here demonstrate that gradient kernel regression can result in models with as good or better performance than that obtained by actually going through the gradient descent training process. 

This result suggests that the model architecture plays a key role in the success of a deep neural network. The kernel representation is constructed from the gradient of the network, and the gradient of the network is a consequence of the network architecture. It has been shown that neural networks with the right architecture, even with random weights, can be used to good effect \cite{he2016powerful,Ulyanov_2020}. It seems plausible that the gradient kernel gets its explanatory power solely from the architecture of the network.

The results also provide a mechanism for testing the performance of a network without going through
the SGD training process. This side steps a number of complexities that arise when training using SGD. Learning rate selection and scheduling, stopping rules, and lack of convergence all go away when using a simple linear regression based on the gradient kernel.

\bibliographystyle{plain} 
\bibliography{main}

\end{document}